\documentclass{article}

\usepackage[preprint]{neurips_2022}

\usepackage[utf8]{inputenc} 
\usepackage[T1]{fontenc}    
\usepackage{hyperref}       
\usepackage{url}            
\usepackage{booktabs}       
\usepackage{amsfonts}       
\usepackage{nicefrac}       
\usepackage{microtype}      
\usepackage{xcolor}         

\usepackage{amsmath}
\usepackage{amsthm}
\usepackage{amssymb}
\usepackage{caption}
\usepackage{subcaption}
\usepackage{enumitem}
\usepackage{graphicx}

\usepackage{natbib}
\newtheorem{assump}{Assumption}

\newtheorem{prop}{Proposition}

\newcommand{\R}{\mathbb{R}}
\newcommand{\C}{\mathbb{C}}
\renewcommand{\i}{\boldsymbol{i}}
\renewcommand{\S}{\mathbb{S}}
\newcommand{\E}{\mathbb{E}}
\newcommand{\V}{\mathbb{V}}
\newcommand{\y}{\boldsymbol{y}}
\newcommand{\X}{\boldsymbol{X}}
\newcommand{\x}{\boldsymbol{x}}
\newcommand{\Z}{\boldsymbol{Z}}

\renewcommand{\a}{\boldsymbol{a}}
\newcommand{\bb}{\boldsymbol{\beta}}
\newcommand{\bs}{\boldsymbol{\sigma}}
\newcommand{\bS}{\boldsymbol{\Sigma}}
\newcommand{\bt}{\boldsymbol{\theta}}
\newcommand{\bT}{\boldsymbol{\Theta}}

\title{Asymptotics of Bayesian Uncertainty Estimation in Random Features Regression}

\author{
    Youngsoo Baek \\
    Department of Statistical Science\\
    Duke University\\
    Durham, NC 27705\\
    \texttt{youngsoo.baek@duke.edu} \\
    \And
    Samuel I. Berchuck \\
    Department of Biostatistics \& Bioinformatics \\
    Duke University\\
    Durham, NC 27705\\
    \texttt{sib2@duke.edu} \\
    \And
    Sayan Mukherjee\thanks{Deparments of Mathematics, Computer Science, Biostatistics \& Bioinformatics, Duke University, NC}
    \thanks{Max Planck Institute for Mathematics in the Sciences, Leipzig} \\
    Center for Scalable Data Analysis and Artificial Intelligence \\
    Universit\"at Leipzig\\
    Leipzig 04105\\
    \texttt{sayan.mukherjee@mis.mpg.de}
}

\begin{document}

\maketitle

\begin{abstract}
   In this paper, we compare and contrast the behavior of the posterior predictive distribution to the risk of the 
maximum a posteriori (MAP) estimator for the random features regression model in the overparameterized regime.  We will focus on the variance of the  posterior predictive distribution (Bayesian model average) and compare its asymptotics to that of the risk of the MAP estimator. In the regime where the model dimensions grow faster than any constant multiple of the number of samples,
asymptotic agreement between these two quantities is governed by the phase transition in the signal-to-noise ratio. They also asymptotically agree with each other when the number of samples grows faster than any constant multiple of model dimensions. Numerical simulations illustrate finer distributional properties of the two quantities for finite dimensions. 
We conjecture they have Gaussian fluctuations and exhibit similar properties as found by previous authors in a Gaussian sequence model, which is of independent theoretical interest.
\end{abstract}


\section{Introduction} \label{section1}

One of the most surprising empirical observations in deep learning is the generalization of overparameterized models that can perfectly interpolate the data. The ``double descent'' curve, referring to the test error first increasing then decreasing with model complexity, has been both empirically and theoretically validated for linear \citep{hastie2022surprises} and nonlinear models \citep{ghorbani2021linearized,mei2022generalization,hu2022sharp}. 
\cite{mei2022generalization} showed that the generalization error of the random features (RF) model proposed by \cite{rahimi2007random} does demonstrate double descent. 
Perhaps more surprisingly, they also showed that vanishingly small regularization can yield optimal generalization in a nearly noiseless learning task. These findings highlight the recent trend of explaining the success of machine learning through the prism of beneficial interpolation and overparameterization \citep{belkin_2021}.

There exists, however, another approach to overparameterized learning problems, which is the Bayesian posterior predictive distribution. In Bayesian practice, one can define the training objective as the negative log-likelihood of a probabilistic model. The ridge regularized predictor for the RF model studied by \cite{mei2022generalization} is the maximum a posteriori (MAP) estimator of this probabilistic model. A ``truly Bayesian'' approach, on the other hand, is to derive the posterior predictive distribution which averages over different predictors and summarizes one's uncertainty in prediction. The posterior predictive distribution is also referred to as the ``Bayesian model average'' in the literature \citep{ovadia2019can,fortuin2021bayesian}. A fundamental question in Bayesian statistics is whether Bayesian credible sets, defined as high probability regions of the posterior, are also valid confidence sets in the frequentist sense \citep{kleijn2012bernstein}. While this is true in finite-dimensional settings and known as Bernstein von-Mises theorem, the behavior of ``frequentist'' and ``Bayesian'' confidence sets can be drastically different in high-to-infinite dimensions, as revealed by \cite{cox1993analysis,freedman1999wald,johnstone2010high}. 

In this work, we examine the frequentist properties of the Bayesian random features model.
 The focus is on whether the variance of the posterior predictive distribution (PPV) demonstrates similar asymptotics as the risk of the MAP estimator studied by \cite{mei2022generalization}. 
 The simplicity of the probabilistic model simplifies the study, as the posterior predictive distribution is available in closed form and is centered around the MAP estimator, for which we know the asymptotic risk. Due to this simplicity, 
 agreement between the risk of the MAP estimator versus the PPV directly implies good coverage property of a natural confidence set formed by the posterior predictive. However, in light of previous work of \cite{johnstone2010high}, it is to be expected that such agreement can be hard to reach in high-dimensional learning. Previous literature relevant to our problem is reviewed in Section \ref{section2_3}.

In this work, we show:
\begin{enumerate}
    \item Absence of double descent phenomenon in the PPV with vanishing regularization (vague prior). Numerical instability does arise at the ``interpolation boundary'' for any finite dimensions.
    \item In low noise problems, the expected width of the Bayesian credible ball for highly overparameterized models can be much wider than the true risk of the MAP predictor.
    \item The expected width of a credible ball asymptotically agrees with the risk when the sample size grows much faster than the model dimensions. Numerical results show the required growth rate of the sample size is unreasonable for many learning problems.
    \item Numerical simulations demonstrate finer distributional properties that are currently beyond the reach of theoretical analysis. They suggest the results of \cite{freedman1999wald} and \cite{johnstone2010high} in the Gaussian sequence model are applicable to the more complicated RF model.
\end{enumerate}

The rest of the paper is organized as follows. Section \ref{section2} reviews the basic setup of the RF model and its probabilistic reformulation. Section \ref{section3} summarizes our main results based on asymptotic formulae and numerical simulations. Section \ref{section4} concludes with a discussion of the implications and future directions.


\section{Background on Random Features Model} \label{section2}

Let inputs $\x_i\in\R^d,\; i = 1,2,\ldots,n$ be drawn i.i.d. from a uniform measure (denoted $\tau$) on a $ d$-dimensional sphere with respect to the conventional Euclidean norm:
\begin{equation}
    \S^{d-1}(\sqrt{d}) := \{\x\in\R^{d}:||\x|| = \sqrt{d}\}.
\end{equation}
Let outputs $y_i$ be generated by the following model:
\begin{equation}
    y_i = f_d(\x_i) + \epsilon_i,\; f_d(\x) = \beta_{d,0} + \langle\x,\bb_d\rangle + f_{d}^{NL}(\x).
    \label{eqn:generating}
\end{equation}
The model is decomposed into a linear component and a nonlinear component, $f_d^{NL}$. The random error $\epsilon_i$'s are assumed to be i.i.d. with mean zero, variance $\tau^2$, and finite fourth moment. We allow $\tau^2=0$, in which case the generating model is \emph{noiseless}. In the analysis, these quantities will be assumed to obey further conditions, including an appropriate scaling so that all $d$-dependent quantities are $O_d(1)$.

\subsection{Training with Ridge Regularization} \label{section2_1}

We focus on learning the optimal \emph{random features model} that best fits the training data. This is a class of functions
\begin{equation}
    \mathcal{F} := \left\{f: f(\x) = \sum_{j=1}^{N} a_j\sigma(\langle\x,\bt_j\rangle/\sqrt{d})\right\},
    \label{eqn:rf_class}
\end{equation}
which is dependent on $N$ random features, $(\bt_j)_{j=1}^N$, which are drawn i.i.d.  from $\mathbb{S}^{d-1}$, and a nonlinear activation function, $\sigma$. The training objective solves a regularized least squares problem for the linear coefficients $\a\equiv(a_j)_{j=1}^{N}$:
\begin{equation}
    {\min}_{\a\in\R^N} \left\{\sum_{i=1}^{n}\left(y_i - \sum_{j=1}^{N} a_j\sigma(\langle \x_i,\theta_j\rangle/\sqrt{d})\right)^2 + 
    d\psi_{1,d}\psi_{2,d}\lambda||\a||^2\right\},
    \label{eqn:ridgep}
\end{equation}
where $\psi_{1,d} = N/d$, $\psi_{2,d} = n/d$, and $\lambda > 0$. The optimal weights, $\widehat{\a}\equiv\widehat{\a}(\lambda)$, determine an optimal ridge predictor denoted by $\widehat{f}\equiv f(\cdot;\widehat{\a}(\lambda))$. The dependence of the trained predictor on the dataset $(\X,\y)$ and features $\bT$ are suppressed in notation.

There exist both practical and theoretical motivations for studying RF regression. On the practical side, RF regression has been suggested by \cite{rahimi2007random} as a randomized approximation scheme for training kernel ridge regression (KRR) models for large datasets. On the theoretical side, \eqref{eqn:ridgep} describes a ``lazy training'' algorithm for a 2-layer neural network with activation function $\sigma$. Previous studies have focused on the approximation power of such a function class \citep{jacot2018neural,chizat2019lazy} and the optimization landscape involved in training problems of type \eqref{eqn:ridgep} \citep{mei2018mean,mei2019mean}.

Given a new input feature $\x$, the ridge regularized predictor for the unknown output $y$ has the form
\begin{equation}
    \widehat{f}(\x) \equiv f({\x};\widehat{\a}) = \sigma({\x}^T\bT/\sqrt{d})\widehat{\a},
    \label{eqn:pred}
\end{equation}
with the optimal ridge weights $\widehat{\a}$ and a resolvent matrix that defines the joint posterior of the weights:
\begin{align}
    \widehat{\a} &\equiv \widehat{\a}(\lambda) := \widehat{\bS}(\lambda){\Z}^T{\y}/\sqrt{d}
    \label{eqn:ridges}, \\
    \widehat{\bS}(\lambda) &:= \left( {\Z}^T{\Z} + \psi_{1d}\psi_{2d}\lambda\mathbf{I}_N \right)^{-1}.
    \label{eqn:ridgecov}
\end{align}
Here, $\Z:=\sigma(\X\bT/\sqrt{d})/\sqrt{d}$ for input design matrix $\X\in\R^{n\times d}$ and output vector $\y\equiv(y_i)_{i=1}^{n}$. Similarly we write $\bs(\x):=\sigma({\x}^T\bT/\sqrt{d})$. The $L^2\equiv L^2(\S^{d-1}(\sqrt{d});\tau)$ generalization error of predictor $\widehat{f}$ is defined by
\begin{equation}
    R_{RF}(\y,\X,\bT,\lambda) := \E_{\x}||f_d(\x) - \widehat{f}(\x)||^2 \equiv ||f_d - \widehat{f}||_{L^2}^2,
    \label{eqn:ge}
\end{equation}
We emphasize that \eqref{eqn:ge} is a random quantity, as it depends on $(\y,\X,\bT)$.

\subsection{RF as Bayesian Model} \label{section2_2}

The objective function \eqref{eqn:ridgep} can be interpreted as a MAP estimation problem for an equivalent Bayesian model. Formally, we adopt a $d$-dependent weight prior distribution, denoted $p(\a)$, and also a normal likelihood, denoted $p(\y|\X,\bT,\a)$, centered around a function in the class \eqref{eqn:rf_class} with variance $\phi^{-1}$.
\begin{align*}
    \a &\sim \mathrm{Normal}\left(0,\phi^{-1}\frac{\psi_{1,d}\psi_{2,d}\lambda}{d}\mathbf{I}_N\right)\\
    \y \mid \X,\bT,\a &\sim \mathrm{Normal}\left(\sigma(\X\bT/\sqrt{d})\a,\phi^{-1}\mathbf{I}_n\right)
\end{align*}
The normal likelihood model need not agree with the generating process \eqref{eqn:generating}, as is often the case for Bayesian deep learning. Technically and inferentially, it can be justified as a coherent update of belief given a squared error loss \citep{bissiri2016general}. The joint likelihood of $(\y,\a)$ is defined conditional on both the random features $\bT$ and an ``inverse temperature'' $\phi$. The choice to condition on $\bT$ instead of learning them can be unnatural in certain settings and is mainly for the convenience of analysis. However, we do note that the RF model has been used by \cite{crawford2018} as an approximation of fully Bayesian kernel learning.

The posterior predictive distribution, or Bayesian model average over the posterior of $\a$, is defined as
\begin{equation}
        p(y \mid \x,\y,\X,\bT) := \int_{\R^N} p(y \mid \x,\bT,\a)~p(\a \mid \y,\X,\bT)~d\a,
\end{equation}
where the posterior of $\a$ is a probability distribution placing higher mass near settings of $\a$ minimizing ridge objective \eqref{eqn:ridgep}:
\begin{equation}
    p(\a \mid \y,\X,\bT) \propto
    \exp\left[-\frac{\phi}{2}
    \left\{\sum_{i=1}^{n}\left(
    y_i - f(\x;\a)
    \right)^2 + \frac{\psi_{1,d}\psi_{2,d}\lambda}{d}||\a||^2\right\}\right].
\end{equation}
This is a Gaussian measure centered around $\widehat{\a}$ \eqref{eqn:ridges} and covariance matrix $\widehat{\bS}(\lambda)/d$. Thus the posterior predictive at new input $\x$ is also a normal distribution centered around $\widehat{f}$, with variance
\begin{equation}
    s^2(\x) \equiv s^2(\x;\lambda) := \phi^{-1}(1+\bs(\x)^T\widehat{\bS}(\lambda)\bs(\x)/d).
    \label{eqn:ppv}
\end{equation}
We refer to \eqref{eqn:ppv} as the PPV. This quantity dictates the width of the uncertainty interval centered around \eqref{eqn:ridgep} evaluated at $\x$. The expected PPV over $\x$ is defined by
\begin{equation}
    S_{RF}^2(\y,\X,\bT,\lambda) := \E_{\x}[s^2(\x)] \equiv \E_{\x}\left[\V[y \mid \x,\y,\X,\bT]\right].
    \label{eqn:e_ppv}
\end{equation}
The expectation over $\x$, similarly as in \eqref{eqn:ge}, yields the radius of the posterior credible ball for a posterior Gaussian process $f$ centered around the optimal predictor $\widehat{f}$. Furthermore, the Gaussian likelihood model simplifies the expected PPV into a decomposition of this radius and the inverse temperature of the posterior:
\begin{equation}
    S_{RF}^2 = \int||f-\widehat{f}||_{L^2}^2~dp(f \mid \y,\X,\bT) + \phi^{-1},
\end{equation}
where $p(f \mid \y,\X,\bT)$ is the law of Gaussian process $f$ induced by the weight posterior $p(\a \mid \y,\X,\bT)$.
 Thus, both summands in the display depend on the training features and are random. It is worth contrasting this definition with \eqref{eqn:e_ppv}; in the next section, we explain in detail the motivation for comparing the two quantities.

The extra quantity introduced in the Bayesian formulation, $\phi$, governs how much the resulting posterior should concentrate around the ridge predictor \eqref{eqn:ridges}. Since we are interested in the regime where $N,d\to\infty$, it is reasonable to assume the scale of the likelihood, $\phi$, must appropriately decrease with $d$, similar to how the prior on $\a$ is rescaled by $1/d$. Practitioners may adopt some prior distribution on this parameter and perform hierarchical inference. We adopt a simpler, empirical Bayes approach and choose it to maximize the marginal likelihood:
\begin{equation}
    \widehat{\phi}^{-1}\equiv \widehat{\phi}^{-1}(\lambda) := \frac{\langle\y,
    \y - \widehat{f}(\X)\rangle}{n}.
    \label{eqn:s_eb}
\end{equation}
This choice coincides with the training set error attained by predictor \eqref{eqn:ridgep}, so it will be decreasing as $N\to\infty$. If $N>n$ and $\lambda\to 0^+$, the training error vanishes as the model can perfectly interpolate the training set. We note the precise asymptotics of the training error has been already characterized by \cite{mei2022generalization} (Section 6).

A fundamental question here is whether $R_{RF}$ and $S^2_{RF}$ have similar asymptotics, as both quantities summarize uncertainty about our prediction in different ways. $R_{RF}$ is the ``frequentist's true risk,'' which requires assumptions about the unknown generative process. $S^2_{RF}$, on the other hand, is the ``Bayesian's risk'' that can be actually computed without any model assumptions. Its asymptotics \emph{does} depend on model assumptions, as both the prior and likelihood need not capture the generative process \eqref{eqn:generating}. In particular, it is desired that it agrees with $R_{RF}$ in some limiting sense. Throughout the rest of this work, we probe the question: Do $R_{RF}$ and $S^2_{RF}$ converge to the same value as $d,n,N\to\infty$?

\subsection{Previous Works} \label{section2_3}

We have introduced our problem of comparing two different quantities, $R_{RF}$ and $S^2_{RF}$, and provided their respective interpretations. Such comparison, between the frequentist risk and the variance of the posterior, was done by \cite{freedman1999wald} in a white noise model, where one observes an infinite square-summable sequence with Gaussian noise. The key finding is that the distributions of two corresponding statistics, re-normalized, have different variances. They are in fact radically different distributions in that they are essentially orthogonal. \cite{knapik2011bayesian} clarified the situation by showing frequentist coverage of credible ball depends heavily on the smoothness of the prior covariance operator. \cite{johnstone2010high} have extended the results to a sequence of finite but growing length, which is a setup more similar to ours. Our study for the RF model addresses a much simpler question, as the theoretical results in Section \ref{section3} only address whether two statistics converge to the same limits. Unlike in the work by previous authors, the identity of limits cannot be taken for granted. A key feature driving the different asymptotics of the two quantities is that the Bayesian's prior and the likelihood are ``mismatched'' with respect to the data generating process; i.e., $f_d$ need not belong to the RF class \eqref{eqn:rf_class}. In Section \ref{section3_3}, we demonstrate some distributional properties of $R_{RF}$ and $S^2_{RF}$ empirically observed in numerical simulations.

Uncertainty quantification in Bayesian learning has recently garnered much attention in the theoretical deep learning literature. We highlight, among many works: \cite{clarte2023theoretical}, characterizing exact asymptotic calibration of a Bayes optimal classifier and a classifier obtained from the generalized approximate message passing algorithm, and \cite{clarte2023double}, demonstrating double descent-like behavior of the Bayes calibration curve in RF models. In particular, quantities $R_{RF}$ and $S^2_{RF}$ studied by our work can be related, respectively, to the ``Bayes-optimal classifier'' and ``empirical risk classifier'' of the latter work (Section 2). The recent work of \cite{guionnet-ko}, on the other hand, explicitly addresses the model mismatch in Bayesian inference by deriving the exact, universal asymptotic formula for generalization of a misspecified Bayes estimator in a rank-1 signal detection problem.


\section{Results} \label{section3}

\subsection{Asymptotic Characterization} \label{section3_1}
We present the main results and illustrate them through numerical simulations. We operate under assumptions identical to those of  \cite{mei2022generalization}, stated below. The proofs are collected in the Supplementary Materials.

\begin{assump}
Define $\psi_{1,d}=N/d$ and $\psi_{2,d}=n/d$. We assume $\psi_{1,d}\to\psi_1<+\infty$ and $\psi_{2,d}\to\psi_2<+\infty$ as $d\to\infty$.
\label{assump1}
\end{assump}

\begin{assump}
Let $\sigma:\R\to\R$ be weakly differentiable and $|\sigma(x)|,|\sigma'(x)|<c_0e^{c_1|x|}$ for some $c_0,c_1<+\infty$. For $G\sim\mathrm{Normal}(0,1)$, the coefficients:
\begin{equation}
    \mu_0 = \E[\sigma(G)],\; \mu_1 = \E[G\sigma(G)],\; \mu_*^2 = E[\sigma^2(G)] - (\mu_0^2 + \mu_1^2)
\end{equation}
are assumed to be greater than 0 (ruling out a linear $\sigma$) and finite.
\label{assump2}
\end{assump}

\begin{assump}
The generating model \eqref{eqn:generating} satisfies
\begin{equation}
    \beta_{0,d}^2\to F_0^2<+\infty,\; ||\bb_{d}||^2\to F_1^2<+\infty;
\end{equation}
furthermore, $f_{NL,d}(\cdot)$ is a centered Gaussian process on $\S^{d-1}(\sqrt{d})$, whose covariance function has the form:
\begin{equation}
    \E[f_{NL,d}(\x_1)f_{NL,d}(\x_2)] = \Sigma_d(\langle\x_1,\x_2\rangle/d).
\end{equation}
The kernel $\Sigma_d$ satisfies
\begin{equation}
    \E[\Sigma_d(x_{(1)}/\sqrt{d})] = 0,\;
    \E[x_{(1)}\Sigma_d(x_{(1)}/\sqrt{d})] = 0\quad
    \text{and}\quad\Sigma_d(1) \to F_*^2<+\infty,
\end{equation}
where $x_{(1)}$ is the first entry of $\x\sim{\rm Unif}(\S^{d-1}(\sqrt{d}))$. The signal-to-noise ratio (SNR) of the model is defined by
\begin{equation}
    \rho = \frac{F_1^2}{F_*^2 + \tau^2}. \label{eqn:snr}
\end{equation}
\label{assump3}
\end{assump}

In this ``linear proportional'' asymptotic regime, we derive an asymptotic formula for the expected PPV akin to that of \cite{mei2022generalization} for the generalization error. Analysis shows Definition 1 of \cite{mei2022generalization}, characterizing the asymptotics of $R_{RF}$, can be straightforwardly applied to also characterize the asymptotics of $S^2_{RF}$.

\begin{prop}
    Denote by $\C_+$ the upper half complex plane: $\{a+b\i:\i=\sqrt{-1},\; b > 0\}$. Let functions $\nu_1,\nu_2:\C_+\to\C_+$ be uniquely defined by the conditions: on $\C_+$, $\nu_1(\xi),\nu_2(\xi)$ are analytic and uniquely satisfy the equations
    \begin{align}
        \nu_1 &= \psi_1\left(-\xi-\nu_2-\frac{\zeta^2\nu_2}{1-\zeta^2\nu_1\nu_2}\right)^{-1},
        \label{eqn:fpe1}\\
        \nu_2 &= \psi_2\left(-\xi-\nu_1-\frac{\zeta^2\nu_1}{1-\zeta^2\nu_1\nu_2}\right)^{-1},
        \label{eqn:fpe2}
    \end{align}
    when $|\nu_1(\xi)|\leq\psi_1/{\rm Im}(\xi)$ and $|\nu_2(\xi)|\leq\psi_2/{\rm Im}(\xi)$, with ${\rm Im}(\xi) > C$ for sufficiently large constant $C$. Define
    \begin{equation}
        \chi = \nu_1(\i\sqrt{\psi_1\psi_2\lambda}/\mu_*)\nu_2(\i\sqrt{\psi_1\psi_2\lambda}/\mu_*),\;
        \zeta = \frac{\mu_1}{\mu_*}.
    \end{equation}
    Under Assumptions \ref{assump1}-\ref{assump3} and for $\lambda > 0$,
    \begin{align*}
    S_{RF}^2(\y,\X,\bT,\lambda)\stackrel{P}{\to}
    \mathcal{S}^2 = \frac{F_1^2}{1-\chi\zeta^2} + F_*^2 + \tau^2~\text{as }d,n,N\to\infty.
    \end{align*}
    \label{prop1}
\end{prop}

Note that $\mathcal{S}^2$ depends on $(\psi_1,\psi_2,\lambda)$ only through function $\chi$. The following facts follow from this fact and the asymptotic characterization of function $\chi$ when $\lambda\to 0^+$.
\begin{prop}
    The following holds for the map $(\psi_1,\psi_2,\lambda)\mapsto\mathcal{S}^2$ defined in Proposition \ref{prop1}:
    \begin{enumerate}
        \item It is non-decreasing in $\lambda$.
        \item $\lim_{\lambda\to 0^+}\mathcal{S}^2<+\infty$ when $\psi_1 = \psi_2$.
    \end{enumerate}
    \label{prop2}
\end{prop}

Item 2, Proposition \ref{prop2} deserves special attention, as it seems to suggest that there is no ``double descent'' in the asymptotics of $S^2$ when using an improper prior. This fact does \emph{not} possess an operational meaning at any finite $N=n$, due to the ordering of the limits $N,n\to\infty$ and $\lambda\to0^+$. In fact, one may expect that the distribution of the least singular value of the relevant matrix $\Z$ causes numerical instability for small $\lambda$ when $N = n$. In Section \ref{section3_3}, this hypothesis is empirically validated through simulations. Subtly, the theoretical prediction does accurately describe the numerical simulations for small choices of $\lambda$. On the other hand, the asymptotic prediction for the frequentist risk does diverge when $\lambda\to 0^+$ and $\psi_1 = \psi_2$.

\subsection{Comparison with Generalization Error} \label{section3_2}

The asymptotics of \eqref{eqn:ge}, the ``frequentist'' risk for our comparison, was characterized by \cite{mei2022generalization} through a theorem similar to Proposition \ref{prop1}. Our main interest lies in comparing the risk against the Bayesian variance in two limiting cases:
\begin{enumerate}
    \item Highly overparameterized models where $\psi_1\to\infty$. The number of parameters grows faster than any constant multiple of the number of samples.
    \item Large sample models where $\psi_2\to\infty$. The number of samples grows faster than any constant multiple of the number of parameters.
\end{enumerate}
The first regime has been studied as the relevant regime in modern deep learning. The second regime is closer to the classical regime of asymptotic statistics where only the number of samples $n$ diverges. Below, we re-state the asymptotic formulae for $R_{RF}$ of \cite{mei2022generalization} in these special cases, which admits simplifications relative to when both $\psi_1,\psi_2$ are finite.

\begin{prop}
\emph{(Theorem 4-5, \cite{mei2022generalization})} Under the notation of Proposition \ref{prop1}, define a function 
\begin{align*}
    \omega\equiv \omega(\zeta,\psi,\bar{\lambda}) = -\frac{(\psi\zeta^2-\zeta^2-1) + \sqrt{\psi\zeta^2-\zeta^2-1}}{2(\bar{\lambda}\psi + 1)}.
\end{align*}
In the highly overparameterized regime, where $\psi_1\to\infty$, asymptotic risk is given as
\begin{align*}
    \mathcal{R}_{wide}(\rho,\zeta,\psi_2,\bar{\lambda}) = \lim_{\psi_1\to\infty}\lim_{d\to\infty}\E R_{RF}(\y,\X,\bT,\lambda) = \frac{(F_1^2 + F_*^2 + \tau^2)(\psi_2\rho + \omega_2^2)}{(1+\rho)(\psi_2-2\omega_2\psi_2 + \omega_2^2\psi_2 - \omega_2^2)} + F_*^2
\end{align*}
with $\omega_2 = \omega(\zeta,\psi_2,\lambda/\mu_*^2)$.

In the large sample regime, where $\psi_2\to\infty$, asymptotic risk is given as
\begin{align*}
    \mathcal{R}_{lsamp}(\zeta,\psi_1,\bar{\lambda}) = \lim_{\psi_2\to\infty}\lim_{d\to\infty}\E R_{RF}(\y,\X,\bT,\lambda) = \frac{F_1^2(\psi_1\zeta^2 + \omega_1^2)}{\zeta^2(\psi_1 - 2\omega_1\psi_1 + \omega_1^2\psi_1 - \omega_1^2)} + F_*^2
\end{align*}
with $\omega_1 = \omega(\zeta,\psi_1,\lambda/\mu_*^2)$.
\label{prop3}
\end{prop}

A simple question, mentioned in Section \ref{section2_2}, was whether the two quantities agree. It turns out that in the second regime, at least, the two formulae converge to the same limit, which is the main content of the next Proposition. In the first regime, whether the limits agree is determined by the signal-to-noise ratio $\rho$. \cite{mei2022generalization} showed that if $\rho$ is larger than a certain critical threshold $\rho_*$, which depends on $(\psi_2,\zeta)$, the optimal regularization is achieved by $\lambda\to 0^+$, whereas if $\rho$ is smaller than $\rho_*$, there exists an optimal regularization $\lambda_*$ bounded away from 0. This phase transition also determines the agreement of the risk and the PPV in the limit.

\begin{prop}
    \emph{(Proposition 5.2,\cite{mei2022generalization}+$\alpha$)}
    Define quantities
    \begin{align*}
        \rho_*(\zeta,\psi_2) &= \frac{\omega_{0,2}^2 - \omega_{0,2}}{(1-\psi_2)\omega_{0,2} + \psi_2}, \\
        \omega_{0,2} &= \omega(\zeta,\psi_2,0)
    \end{align*}
    under the notation of Proposition \ref{prop3}.
    
    \begin{enumerate}
        \item If $\rho < \rho_*$, then $\min_{\bar{\lambda}\geq 0}\mathcal{R}_{wide}(\rho,\zeta,\psi_2,\bar{\lambda})$, is attained at some $\lambda^{opt} > 0$:
        \begin{align*}
            \lambda^{opt} &:= {\arg\min}_{\bar{\lambda}\geq 0}\mathcal{R}_{wide}(\rho,\zeta,\psi_2,\bar{\lambda}) =  \frac{\zeta^2\psi_2 - \zeta^2\omega_*\psi_2 + \zeta^2\omega_* + \omega_* - \omega_*^2}{(\omega_*^2-\omega_*)\psi_2}, \\
            \omega_* &:= \omega(\sqrt{\rho},\psi_2,0).
        \end{align*}
        Furthermore, $\mathcal{R}_{wide}(\rho,\zeta,\psi_2,\lambda^{opt}) = \lim_{\psi_1\to\infty} \mathcal{S}^2(\psi_1,\psi_2,\lambda^{opt}) - \tau^2$.
        
        If, on the other hand, $\rho > \rho_*$, then ${\arg\min}_{\bar{\lambda}\geq 0}\mathcal{R}_{wide}(\rho,\zeta,\psi_2,\bar{\lambda})  = 0$. Furthermore, $\mathcal{R}_{wide}(\rho,\zeta,\psi_2,0) < \lim_{\psi_1\to\infty} \mathcal{S}^2(\psi_1,\psi_2,0) - \tau^2$.
        \item ${\arg\min}_{\bar{\lambda}\geq 0}\mathcal{R}_{lsamp}(\zeta,\psi_1,\bar{\lambda}) = 0$ and $\mathcal{R}_{lsamp}(\zeta,\psi_1,0) = \lim_{\psi_2\to\infty} \mathcal{S}^2(\psi_1,\psi_2,0) - \tau^2$.
    \end{enumerate}
    \label{prop4}
\end{prop}

Note the subtraction of the noise level $\tau^2$ from $\mathcal{S}^2$. This is because $S_{RF}^2$ is computed based on the training error in Proposition \ref{prop1}, which includes both the approximation error and the variance of data, whereas $R_{RF}$ is computed based only on the approximation error in the test set.

\subsection{Numerical Simulations} \label{section3_3}

\begin{figure}[!tb]
     \begin{subfigure}{0.48\textwidth}
        \centering
         \includegraphics[width=\textwidth]{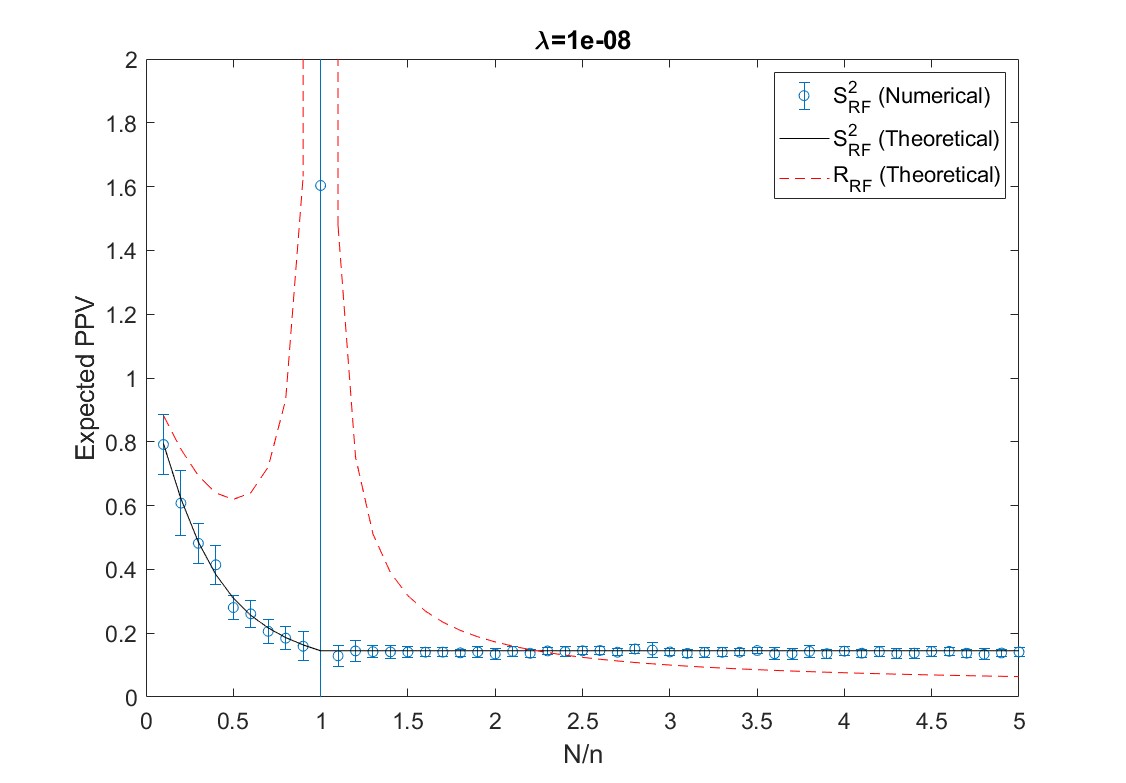}
         \caption{$\lambda=\text{1e-8}$}
         \label{fig:interp_vars}
     \end{subfigure}
     \hfill
     \begin{subfigure}{0.48\textwidth}
        \centering
         \includegraphics[width=\textwidth]{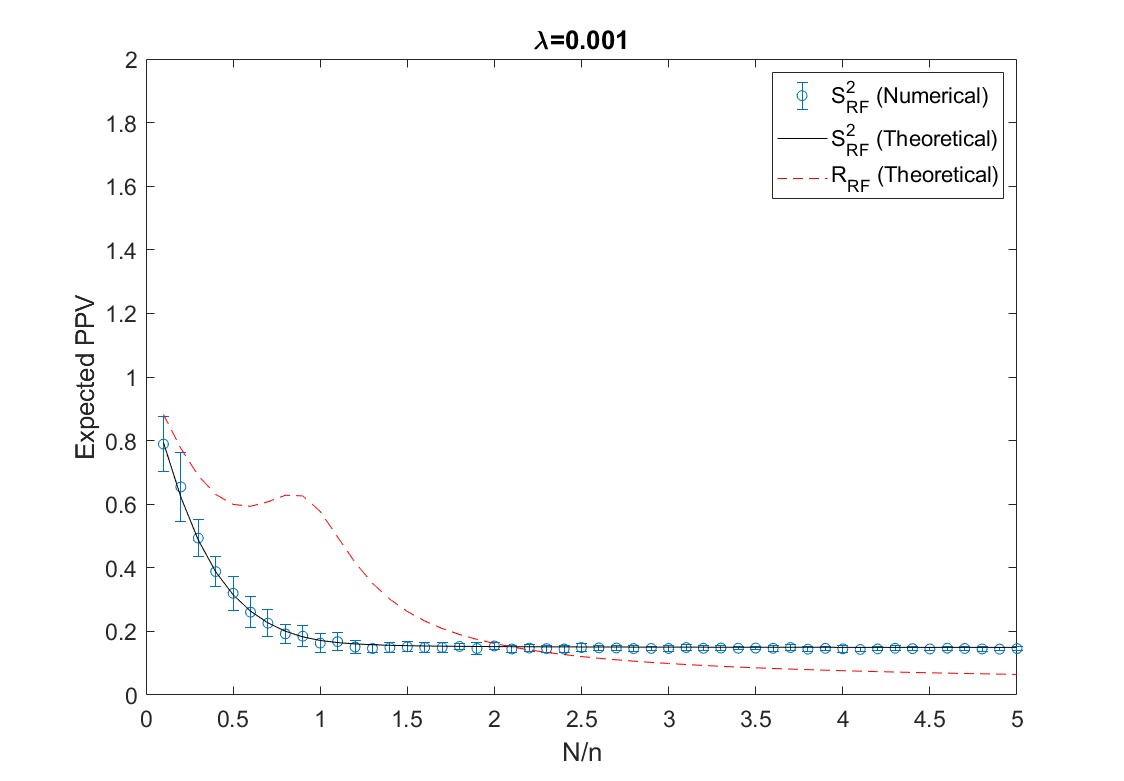}
         \caption{$\lambda=\text{1e-3}$}
         \label{fig:reg_vars}
     \end{subfigure}
        \caption{Comparison of asymptotic formula and 20 instances of $S_{RF}^2$ \eqref{eqn:e_ppv}. Data are generated via noiseless linear model $y=\langle\x,\bb\rangle$ ($||\bb||=1$, $\rho = \infty$). Activation is ReLU: $\sigma(x) = \max\{0,x\}$. $d$ and $n$ are fixed to 100 and 300, respectively. The asymptotic formula for $R_{RF}$ \eqref{eqn:ge} is plotted for comparison ({\color{red} red}, dashed).}
        \label{fig:vars_l}
\end{figure}

\begin{figure}[!tb]
     \begin{subfigure}{0.48\textwidth}
        \centering
         \includegraphics[width=\textwidth]{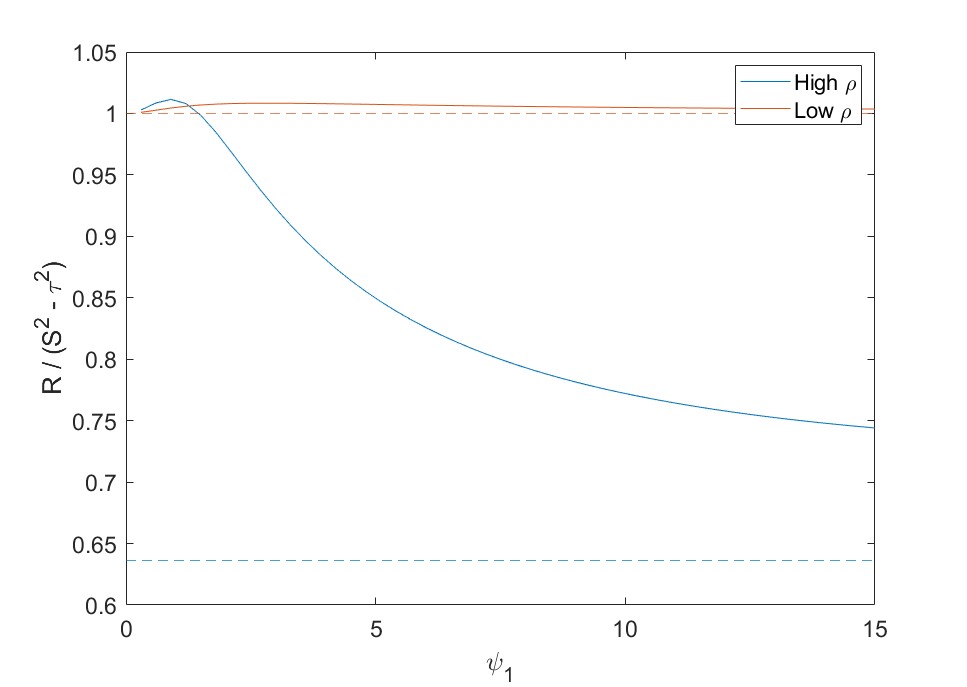}
         \caption{$\psi_2 = 3$}
         \label{fig:tune_psi1}
     \end{subfigure}
     \hfill
     \begin{subfigure}{0.48\textwidth}
        \centering
         \includegraphics[width=\textwidth]{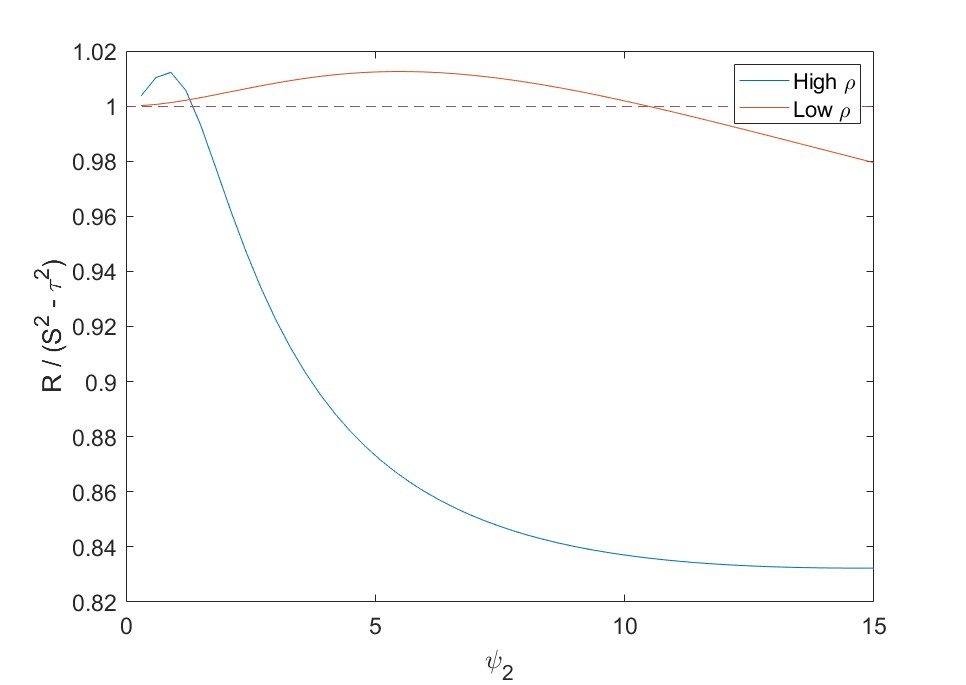}
         \caption{$\psi_1 = 3$}
         \label{fig:tune_psi2}
     \end{subfigure}
    \caption{Ratio of $\mathcal{R}(\lambda^{opt})$ to $\mathcal{S}^2(\lambda^{opt})-\tau^2$ as a function of $\psi_1$ (\ref{fig:tune_psi1}) and of $\psi_2$ (\ref{fig:tune_psi2}). In each plot, $\psi_2$ and $\psi_1$ are respectively fixed to 3, while $F_1 = 1$, $F_* = 0$, and $\rho = 1/\tau^2$ for noise variance $\tau^2\in\{.2, 5\}$.}
    \label{fig:opt}
\end{figure}

\begin{figure}[!tb]
    \centering
    \includegraphics[width=\textwidth]{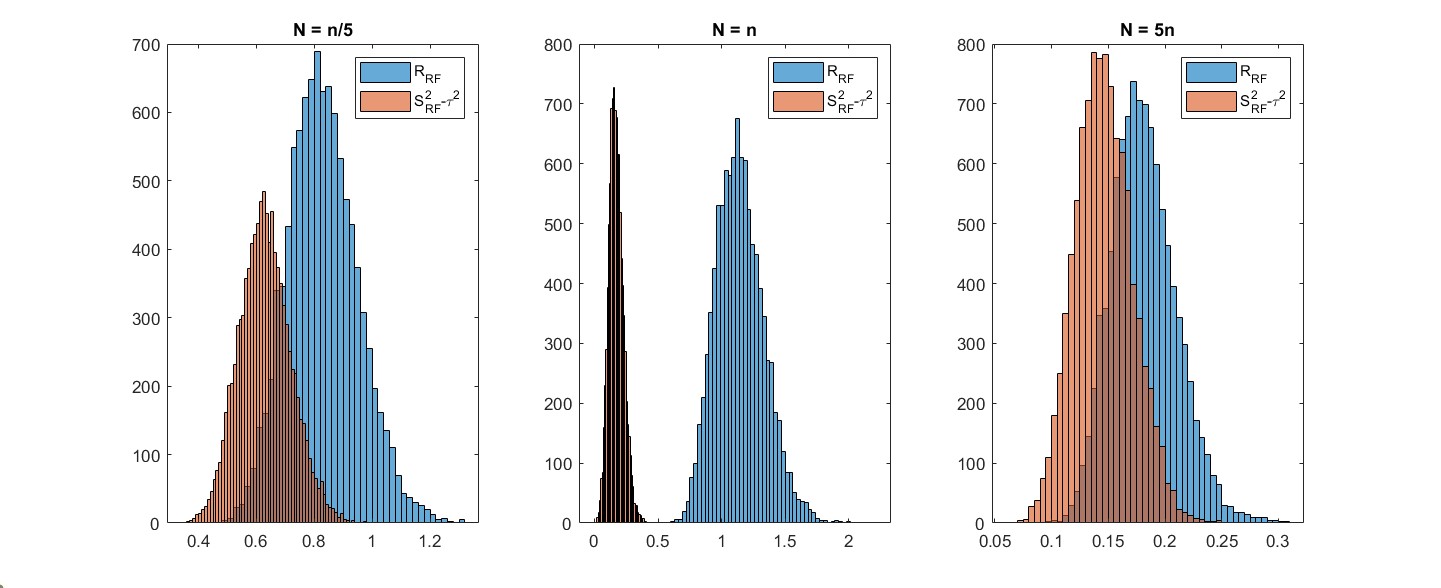}
    \caption{Histograms of 1e+4 draws of $R_{RF}$ and $S^2_{RF}-\tau^2$ under low-noise linear model $y=\langle\x,\bb\rangle + \tau^2$ with $\tau^2 = 1/5$.}
    \label{fig:distn}
\end{figure}

In this section, we first compare evaluations of the asymptotic formulae for $R_{RF}$ and $S^2_{RF}$ for varying $(\psi_1,\psi_2,\lambda)$. We highlight their difference for finite $(\psi_1,\psi_2)$ at the optimal choice of $\lambda$ for the MAP risk. We also present numerical simulations, which both validate the formulae (they concentrate fast) and empirically exhibit further interesting distributional properties (suggesting the need for second-order asymptotics). 

Figure \ref{fig:vars_l} shows the dramatic mismatch between $R_{RF}$ and $S^2_{RF}$ in the low-noise regime. It turns out the conservativeness of the credible ball persists for the choice of $\lambda$ that depends on $(\psi_1,\psi_2)$. The situation becomes more delicate in the high-noise regime because there exists a phase transition in the optimal choice of $\lambda$ that depends on \eqref{eqn:snr}, which decreases with the noise variance $\tau^2$. Figures \ref{fig:opt}.\ref{fig:tune_psi1} and \ref{fig:tune_psi2} 
compare the two curves, $\mathcal{R}$ and $\mathcal{S}$, for the ``optimally tuned'' $\lambda$ at which the best possible frequentist risk is attained, for a fixed pair of $(\psi_1,\psi_2)$. In a low-noise task, with $\rho = 5$, the ratio of the frequentist risk of the posterior mean predictor to the width of the posterior predictive credible ball is less than 1 for a wide range of $\psi_1$. The situation is more nuanced and possibly more favorable for the Bayesian case in the high-noise task with $\rho = 1/5$. 

While Figure \ref{fig:vars_l} validates good concentration properties of $R_{RF}$ and $S^2_{RF}$ with respect to their asymptotic formulae, we may want to investigate the rate of fluctuations for these quantities. Figure \ref{fig:distn} suggests an interesting phenomenon: both quantities appear Gaussian, but are nearly orthogonal precisely near the ``interpolation boundary'' where $R_{RF}$ exhibits double descent (2nd subplot). Our empirical results should be compared with the results of \cite{freedman1999wald} and \cite{johnstone2010high}. The Gaussianity of $R_{RF}$ and $S^2_{RF}$, if true, strengthens the agreement between the expected width $S^2_{RF}$ and expected risk $R_{RF}$ immediately transfers to the agreement between the frequentist coverage of $(1-\alpha)-\%$ Bayes credible ball around the mean and the nominal coverage. 
Again, such claims are generally true in finite-dimensional settings, but not true for high-dimensional settings. For instance, \cite{freedman1999wald} showed that asymptotically, the posterior variance fluctuates like a Gaussian random variable with a variance strictly smaller than that of the frequentist risk. Extracting such information is possible only if we study \emph{second-order information} of $R_{RF}$ and $S_{RF}^2$; in particular, 
suggests a Gaussian central limit for these quantities.  A rigorous proof of a central limit theorem for $R_{RF}$ and $S_{RF}^2$ goes beyond the scope of this work. Nevertheless, we conjecture the following for the fluctuations of $R_{RF}$ and $S_{RF}^2$:
\begin{enumerate}
    \item Both $d(R_{RF} - \E R_{RF})$ and $d(S^2_{RF}-\E S^2_{RF})$ weakly converge to Gaussian distribution with appropriate variances. The faster convergence rate of $d$ rather than $\sqrt{d}$ is known to be common in central limit theorems for linear spectral statistics \citep{lytova2009central}.
    \item When $\psi_1 \leq \psi_2$, asymptotic variance of $S_{RF}^2$ is smaller than $R_{RF}$. When $\psi_1 = \psi_2$, the two distributions are nearly orthogonal. For large enough $\psi_1$, the asymptotic variances are of the same order.
\end{enumerate}

These conjectures are of independent interest and pose interesting theoretical challenges. We must note that second-order asymptotics has received less attention in the deep learning community. Only recently did \cite{li2021asymptotic} study the asymptotic normality of prediction risk achieved by a min-norm least squares interpolator. Their result relies on a central limit theorem for linear statistics of eigenvalues of large sample covariance matrices, established by \cite{bai2004clt}. Many central limit theorems in random matrix theory seem insufficient to handle kernel learning or learning with random features.


\section{Discussion} \label{section4}

In sum, our calculations and empirical findings suggest that the RF model still has some interesting discoveries to offer, particularly for those interested in the implications of Bayesian deep learning. The asymptotics of the posterior predictive summaries can be very different from that of the generalization error. Numerical simulations suggest that some version of the central limit holds so that depending on the particular scaling of dimensions, the frequentist coverage of the posterior predictive distribution can be arbitrarily low or high. We now conclude with a discussion of several conjectures and the next directions suggested by these findings.

\noindent {\bf Heuristic Implications.} While our technical calculations are only applicable to the RF model, we believe the general perspective of comparing frequentist versus Bayesian estimation approaches to the deep learning model can offer useful heuristics explaining empirically observed phenomena in training Bayesian deep learning models. 

One suggestion is that the different regimes of asymptotics may explain away the ``cold posterior effect'', first empirically observed by \cite{wenzel2020good}.
The authors forcefully suggested that raising the posterior of a deep learning model to an even larger power (i.e., more concentrated) can improve generalization. That the posteriors can be too wide in highly overparameterized, near-noiseless regimes may explain this issue. Even though in this simple RF model setup, the posterior predictive mean is left unchanged from that of the posterior, it is plausible that for more complicated models this need not be so. When that posterior is much wider than the expected risk of the mean, we will mix over too many bad predicting weights, so it will be actually worse to average over the posterior than not.

\noindent {\bf Technical Improvements.} The technically most interesting but quite challenging direction, suggested by our numerical simulations in Section \ref{section3_3}, is to study second-order fluctuations of random matrix quantities often encountered in deep learning. While at this point a vast amount of literature exists that systematically overcomes the non-standard definitions of quantities like generalization error from the viewpoint of random matrix theory, we are aware of no study as of yet that overcomes the same issues to show a central limit-type result. On the other hand, central limit theorems for linear spectral statistics remain an active area of research in random matrix theory and are generally acknowledged to require more advanced, subtler arguments that are fundamentally different from first-order results (i.e., convergence of empirical spectral distribution). For a start, a separate work on the central limit theorem for linear spectral statistics of kernel matrices is in preparation by the authors.

Another interesting direction is to extend the comparison between frequentist and Bayesian modeling approaches to more complex models of training dynamics. For instance, \cite{adlam2020neural} have derived precise asymptotics of generalization error in the neural tangent kernel (NTK) regime, in which the features $\bT$ are ``learned'' through a linearized approximation of the gradient flow in training. \cite{mei2018mean,mei2019mean} have suggested a different asymptotic model, in which $\bT$ evolve through nonlinear dynamics. An interesting challenge, both technically and conceptually, is to adapt their analysis to the Bayesian setting, where the gradient flow on the space of posterior measures of $\bT$ is approximated and specific projections are tracked.

\noindent {\bf Making the ``Right'' Comparison.} A possible objection is that the comparison between two quantities is not appropriate: if interested in the coverage of function $f_d$ with our credible set, then one must compare the expected width of the interval of the function space posterior, $\widehat{f}$, instead of that of the Bayesian model average. While the objection is sensible, the fact is that the average variance of the function posterior will be an even worse uncertainty estimate for underparameterized models, whereas it will make little difference for overparameterized models. The reason is that the variance of the posterior of $\widehat{f}$ only differs from that of the posterior predictive, in \eqref{eqn:ppv}, by the training set error term \eqref{eqn:s_eb}. The training error dominates the PPV precisely for underparameterized models ($\psi_1<\psi_2$), so the curves of Figure \ref{fig:vars_l} will now be non-\emph{decreasing} and asymptotically convergent to a similar constant in expectation. This fact illustrates that a simple interpretation of the width of the posterior is complicated in even simplistic high-dimensional models.

Another possible objection is that the same value of $\lambda$ need not be optimal for the posterior predictive. Since the asymptotic curve for $S^2$ is non-increasing in $\lambda$, this implies one must choose even smaller $\lambda$ than $\lambda^{opt}$ to obtain less conservative credible intervals, regardless of the SNR $\rho$. This seems an odd consequence of using overparameterized models, where a good frequentist coverage is obtained by choosing a narrower prior than that with good ``contraction property'' (i.e., minimal $L^2$ risk of posterior mean).

The final objection is that the RF model is simply not a very realistic model of practical Bayesian deep learning. We point out that quantifying the behavior of RF models as approximations of kernel learning methods remains a valid concern for Bayesian practitioners.

\section*{Acknowledgments}
All authors would like to thank Matthew M. Engelhard, Boyao Li, David Page, and Alexander J Thomson for the helpful discussion around the theoretical work. 

Samuel I. Berchuck would like to acknowledge support from the National Eye Institute of the National Institutes of Health under Award Number K99EY033027. The content is solely the responsibility of the authors and does not necessarily represent the official views of the National Institutes of Health.
Sayan Mukherjee would like to acknowledge partial funding from HFSP RGP005, NSF DMS 17-13012, NSF BCS 1552848, NSF DBI 1661386, NSF IIS 15-46331, NSF DMS 16-13261, as well as high-performance computing partially supported by grant 2016- IDG-1013 from the North Carolina Biotechnology Center as well as the Alexander von Humboldt Foundation, the BMBF and the Saxony State Ministry for Science.

\bibliography{main}

\begin{thebibliography}{27}
\providecommand{\natexlab}[1]{#1}
\providecommand{\url}[1]{\texttt{#1}}
\expandafter\ifx\csname urlstyle\endcsname\relax
  \providecommand{\doi}[1]{doi: #1}\else
  \providecommand{\doi}{doi: \begingroup \urlstyle{rm}\Url}\fi

\bibitem[Hastie et~al.(2022)Hastie, Montanari, Rosset, and
  Tibshirani]{hastie2022surprises}
Trevor Hastie, Andrea Montanari, Saharon Rosset, and Ryan~J Tibshirani.
\newblock Surprises in high-dimensional ridgeless least squares interpolation.
\newblock \emph{The Annals of Statistics}, 50\penalty0 (2):\penalty0 949--986,
  2022.

\bibitem[Ghorbani et~al.(2021)Ghorbani, Mei, Misiakiewicz, and
  Montanari]{ghorbani2021linearized}
Behrooz Ghorbani, Song Mei, Theodor Misiakiewicz, and Andrea Montanari.
\newblock Linearized two-layers neural networks in high dimension.
\newblock \emph{The Annals of Statistics}, 49\penalty0 (2):\penalty0
  1029--1054, 2021.

\bibitem[Mei and Montanari(2022)]{mei2022generalization}
Song Mei and Andrea Montanari.
\newblock The generalization error of random features regression: Precise
  asymptotics and the double descent curve.
\newblock \emph{Communications on Pure and Applied Mathematics}, 75\penalty0
  (4):\penalty0 667--766, 2022.

\bibitem[Hu and Lu(2022)]{hu2022sharp}
Hong Hu and Yue~M Lu.
\newblock Sharp asymptotics of kernel ridge regression beyond the linear
  regime.
\newblock \emph{arXiv preprint arXiv:2205.06798}, 2022.

\bibitem[Rahimi and Recht(2007)]{rahimi2007random}
Ali Rahimi and Benjamin Recht.
\newblock Random features for large-scale kernel machines.
\newblock \emph{Advances in neural information processing systems}, 20, 2007.

\bibitem[Belkin(2021)]{belkin_2021}
Mikhail Belkin.
\newblock Fit without fear: remarkable mathematical phenomena of deep learning
  through the prism of interpolation.
\newblock \emph{Acta Numerica}, 30:\penalty0 203–248, 2021.

\bibitem[Ovadia et~al.(2019)Ovadia, Fertig, Ren, Nado, Sculley, Nowozin,
  Dillon, Lakshminarayanan, and Snoek]{ovadia2019can}
Yaniv Ovadia, Emily Fertig, Jie Ren, Zachary Nado, David Sculley, Sebastian
  Nowozin, Joshua Dillon, Balaji Lakshminarayanan, and Jasper Snoek.
\newblock Can you trust your model's uncertainty? evaluating predictive
  uncertainty under dataset shift.
\newblock \emph{Advances in neural information processing systems}, 32, 2019.

\bibitem[Fortuin et~al.(2021)Fortuin, Garriga-Alonso, Ober, Wenzel, R{\"a}tsch,
  Turner, van~der Wilk, and Aitchison]{fortuin2021bayesian}
Vincent Fortuin, Adri{\`a} Garriga-Alonso, Sebastian~W Ober, Florian Wenzel,
  Gunnar R{\"a}tsch, Richard~E Turner, Mark van~der Wilk, and Laurence
  Aitchison.
\newblock Bayesian neural network priors revisited.
\newblock \emph{arXiv preprint arXiv:2102.06571}, 2021.

\bibitem[Kleijn and van~der Vaart(2012)]{kleijn2012bernstein}
BJK Kleijn and AW~van~der Vaart.
\newblock The bernstein-von-mises theorem under misspecification.
\newblock \emph{Electronic Journal of Statistics}, 6:\penalty0 354--381, 2012.

\bibitem[Cox(1993)]{cox1993analysis}
Dennis~D Cox.
\newblock An analysis of bayesian inference for nonparametric regression.
\newblock \emph{The Annals of Statistics}, pages 903--923, 1993.

\bibitem[Freedman(1999)]{freedman1999wald}
David Freedman.
\newblock Wald lecture: On the bernstein-von mises theorem with
  infinite-dimensional parameters.
\newblock \emph{The Annals of Statistics}, 27\penalty0 (4):\penalty0
  1119--1141, 1999.

\bibitem[Johnstone(2010)]{johnstone2010high}
Iain~M Johnstone.
\newblock High dimensional bernstein-von mises: simple examples.
\newblock \emph{Institute of Mathematical Statistics Collections}, 6:\penalty0
  87, 2010.

\bibitem[Jacot et~al.(2018)Jacot, Gabriel, and Hongler]{jacot2018neural}
Arthur Jacot, Franck Gabriel, and Cl{\'e}ment Hongler.
\newblock Neural tangent kernel: Convergence and generalization in neural
  networks.
\newblock \emph{Advances in neural information processing systems}, 31, 2018.

\bibitem[Chizat et~al.(2019)Chizat, Oyallon, and Bach]{chizat2019lazy}
Lenaic Chizat, Edouard Oyallon, and Francis Bach.
\newblock On lazy training in differentiable programming.
\newblock \emph{Advances in neural information processing systems}, 32, 2019.

\bibitem[Mei et~al.(2018)Mei, Montanari, and Nguyen]{mei2018mean}
Song Mei, Andrea Montanari, and Phan-Minh Nguyen.
\newblock A mean field view of the landscape of two-layer neural networks.
\newblock \emph{Proceedings of the National Academy of Sciences}, 115\penalty0
  (33):\penalty0 E7665--E7671, 2018.

\bibitem[Mei et~al.(2019)Mei, Misiakiewicz, and Montanari]{mei2019mean}
Song Mei, Theodor Misiakiewicz, and Andrea Montanari.
\newblock Mean-field theory of two-layers neural networks: dimension-free
  bounds and kernel limit.
\newblock In \emph{Conference on Learning Theory}, pages 2388--2464. PMLR,
  2019.

\bibitem[Bissiri et~al.(2016)Bissiri, Holmes, and Walker]{bissiri2016general}
Pier~Giovanni Bissiri, Chris~C Holmes, and Stephen~G Walker.
\newblock A general framework for updating belief distributions.
\newblock \emph{Journal of the Royal Statistical Society: Series B (Statistical
  Methodology)}, 78\penalty0 (5):\penalty0 1103--1130, 2016.

\bibitem[Crawford et~al.(2018)Crawford, Wood, Zhou, and
  Mukherjee]{crawford2018}
Lorin Crawford, Kris~C. Wood, Xiang Zhou, and Sayan Mukherjee.
\newblock Bayesian approximate kernel regression with variable selection.
\newblock \emph{Journal of the American Statistical Association}, 113\penalty0
  (524):\penalty0 1710--1721, 2018.

\bibitem[Knapik et~al.(2011)Knapik, van~der Vaart, and van
  Zanten]{knapik2011bayesian}
BT~Knapik, AW~van~der Vaart, and JH~van Zanten.
\newblock Bayesian inverse problems with gaussian priors.
\newblock \emph{The Annals of Statistics}, 39\penalty0 (5):\penalty0
  2626--2657, 2011.

\bibitem[Clart{\'e} et~al.(2023{\natexlab{a}})Clart{\'e}, Loureiro, Krzakala,
  and Zdeborov{\'a}]{clarte2023theoretical}
Lucas Clart{\'e}, Bruno Loureiro, Florent Krzakala, and Lenka Zdeborov{\'a}.
\newblock Theoretical characterization of uncertainty in high-dimensional
  linear classification.
\newblock \emph{Machine Learning: Science and Technology}, 4\penalty0
  (2):\penalty0 025029, 2023{\natexlab{a}}.

\bibitem[Clart{\'e} et~al.(2023{\natexlab{b}})Clart{\'e}, Loureiro, Krzakala,
  and Zdeborov{\'a}]{clarte2023double}
Lucas Clart{\'e}, Bruno Loureiro, Florent Krzakala, and Lenka Zdeborov{\'a}.
\newblock On double-descent in uncertainty quantification in overparametrized
  models.
\newblock In \emph{International Conference on Artificial Intelligence and
  Statistics}, pages 7089--7125. PMLR, 2023{\natexlab{b}}.

\bibitem[Guionnet et~al.(2023)Guionnet, Ko, Krzakala, and
  Zdeborová]{guionnet-ko}
Alice Guionnet, Justin Ko, Florent Krzakala, and Lenka Zdeborová.
\newblock Estimating rank-one matrices with mismatched prior and noise:
  universality and large deviations, 2023.

\bibitem[Lytova and Pastur(2009)]{lytova2009central}
A~Lytova and L~Pastur.
\newblock Central limit theorem for linear eigenvalue statistics of random
  matrices with independent entries.
\newblock \emph{Annals of Probability}, 37\penalty0 (5):\penalty0 1778--1840,
  2009.

\bibitem[Li et~al.(2021)Li, Xie, and Wang]{li2021asymptotic}
Zeng Li, Chuanlong Xie, and Qinwen Wang.
\newblock Asymptotic normality and confidence intervals for prediction risk of
  the min-norm least squares estimator.
\newblock In \emph{International Conference on Machine Learning}, pages
  6533--6542. PMLR, 2021.

\bibitem[Bai and Silverstein(2004)]{bai2004clt}
ZD~Bai and Jack~W Silverstein.
\newblock Clt for linear spectral statistics of large-dimensional sample
  covariance matrices.
\newblock \emph{The Annals of Probability}, 32\penalty0 (1A):\penalty0
  553--605, 2004.

\bibitem[Wenzel et~al.(2020)Wenzel, Roth, Veeling, {\'S}wi{k{a}}tkowski, Tran,
  Mandt, Snoek, Salimans, Jenatton, and Nowozin]{wenzel2020good}
Florian Wenzel, Kevin Roth, Bastiaan~S Veeling, Jakub {\'S}wi{k{a}}tkowski,
  Linh Tran, Stephan Mandt, Jasper Snoek, Tim Salimans, Rodolphe Jenatton, and
  Sebastian Nowozin.
\newblock How good is the bayes posterior in deep neural networks really?
\newblock \emph{arXiv preprint arXiv:2002.02405}, 2020.

\bibitem[Adlam and Pennington(2020)]{adlam2020neural}
Ben Adlam and Jeffrey Pennington.
\newblock The neural tangent kernel in high dimensions: Triple descent and a
  multi-scale theory of generalization.
\newblock In \emph{International Conference on Machine Learning}, pages 74--84.
  PMLR, 2020.

\end{thebibliography}


\clearpage

\appendix

\section{Proofs}

\subsection*{Proof of Proposition \ref{prop1}}
The necessary calculations follow immediately from the proof of \cite{mei2022generalization}. Since we do not wish to replicate the long proof with only minor readjustments, we only sketch the main argument. The expected PPV consists of two random variables, denoted by
\begin{equation}
    \widehat{\phi^{-1}}(f_d,\X,\bT,\lambda) = \frac{\langle\y,\y-\widehat{f}(\x)\rangle}{n},\;
    r(\X,\bT,\lambda) := \E_{\x}[\bs(\x)^T(\Z^T\Z + \lambda\psi_{1d}\psi_{2d}\mathbf{I}_N)^{-1}\bs(\x)/d].
    \label{eqn:limit1}
\end{equation}
It suffices to show the two quantities individually converge in probability to some limits. The first of the two is a training set error, for which an asymptotic formula was derived in Section 6, \cite{mei2022generalization}:
\begin{equation}
    \lim_{d\to\infty}\E \widehat{\phi^{-1}} = -\i\nu_2(\i\sqrt{\psi_1\psi_2\lambda}/\mu_*)\sqrt\frac{\lambda\psi_1}{\psi_2\mu_*^2}
    \left(\frac{F_1^2}{1-\chi\zeta^2} + F_*^2+\tau^2\right),
\end{equation}
using the definition of quantities in Proposition \ref{prop1}, for $\chi\equiv \chi(\i\sqrt{\psi_1\psi_2\lambda}/\mu_*) = \nu_1\nu_2$ ($\i=\sqrt{-1}$). At this point, rearranging the fixed point equations \eqref{eqn:fpe1} and \eqref{eqn:fpe2} with $\xi\equiv\i\sqrt{\psi_1\psi_2\lambda}/\mu_*$, we obtain:
\begin{align}
    -\psi_1-\i\sqrt{\lambda\psi_1\psi_2}/\mu_*\nu_1 = -\psi_2- & 
    \i\sqrt{\lambda\psi_1\psi_2}/\mu_*\nu_2 = 
    -\chi-\frac{\zeta^2\chi}{1-\zeta^2\chi}.\\
    \implies
    -\i\sqrt{\frac{\lambda\psi_1}{\psi_2\mu_*^2}}\nu_2 &= 1 + \frac{\chi}{\psi_2}\left(\frac{\zeta^2}{1-\zeta^2\chi} + 1\right).
    \label{eqn:rearranged}
\end{align}

A similar formula for $\E r$ can also be derived, using almost exactly the same steps required for simplifying asymptotic formulae of $R_{RF}$ and $\widehat{\phi^{-1}}$. Assumption \ref{assump3} on the nonlinear component of the generating model, $f_{d}^{NL}$, implies that it admits an explicit decomposition into spherical harmonics that form a basis on $\S^{d-1}$ and coefficients with Gaussian distributions (appropriately scaled). With many simplifications, also sketched out in Appendix E, \cite{mei2022generalization}, we can write
\begin{equation}
    \E r = \mu_1^2{\rm Tr}((\Z^T\Z + \lambda\psi_1\psi_2\mathbf{I}_N)^{-1}(\bT\bT^T/d)) + \mu_*^2{\rm Tr}((\Z^T\Z + \lambda\psi_1\psi_2\mathbf{I}_N)^{-1}) + o_d(1).
\end{equation}

Next, form an $(N+n)\times(N+n)$-dimensional block matrix $\boldsymbol{A}$, indexed by a 5-d real parameter $\boldsymbol{q}=(s_1,s_2,t_1,t_2,p)$:
\begin{equation}
    \boldsymbol{A}(\boldsymbol{q}) = 
    \begin{bmatrix}
    s_1\mathbf{I}_N + s_2\bT\bT/d & \Z^T + p\bT\X^T/\sqrt{d} \\
    \Z + p\X\bT^T/\sqrt{d} & t_2\mathbf{I}_n + t_2\X\X/d
    \end{bmatrix}.
\end{equation}
Denoting by $\xi\in\C_+$ the log-determinant of the resolvent matrix $\boldsymbol{A}(\boldsymbol{q}) - \xi\mathbf{I}_{N+n}$ as $G_d(\xi;\boldsymbol{q})$, straightforward calculus and plugging in the figures reveals
\begin{equation}
    \E r = -\mu_1^2\cdot \i\frac{1}{\sqrt{\psi_1\psi_2\lambda}}\E[\partial_{s_1}G_d(\i\sqrt{\lambda\psi_1\psi_2};0)] -
    \mu_*^2\cdot \i\frac{1}{\sqrt{\psi_1\psi_2\lambda}}\E[\partial_{s_2} G_d(\i\sqrt{\lambda\psi_1\psi_2};0)] + o_d(1).
\end{equation}
At this point, we pass on to the limit $d\to\infty$ and plug in appropriate asymptotic limits of the partial derivatives. Since the $\xi$-derivative of $G_d$ is the trace of $\boldsymbol{A}(\boldsymbol{q}) - \xi\mathbf{I}_{N+n}$, this is possible by characterizing the asymptotics of the Stieltjes transform of the empirical spectral distribution of $\boldsymbol{A}(\boldsymbol{q})$, which is done using Proposition 8.3, \cite{mei2022generalization}. Integrating the limit over path stretching in the imaginary axis ($\i K$ with $K\to\infty$), one then rigorously passes onto the limit by Proposition 8.4, \cite{mei2022generalization}. Final simplifications using Lemma 8.1, \cite{mei2022generalization} yield the formula
\begin{equation}
    \lim_{d\to\infty} \E r = -\i\nu_1(\i\sqrt{\psi_1\psi_2\lambda}/\mu_*)\frac{\mu_*}{\sqrt{\lambda\psi_1\psi_2}}\left(\frac{\zeta^2}{1-\zeta^2\chi} + 1\right).
    \label{eqn:limit2}
\end{equation}
At this point, we may conclude, using \eqref{eqn:limit1},\eqref{eqn:limit2} and \eqref{eqn:rearranged}:
\begin{align*}
    \widehat{\phi^{-1}}(1 + r) &\stackrel{P}{\to} \frac{F_1^2}{1-\zeta^2\chi} + F_*^2 + \tau^2.
\end{align*}

\subsection*{Proof of Proposition \ref{prop2}}
The first point is a consequence of the fact that $S_{RF}^2$ is decreasing in $\lambda$ and passing onto the limits. The second point is immediately seen by taking the limit of $\chi$ as $\lambda\to 0$, a formula for which is provided by Theorem 3, \cite{mei2022generalization}.

\subsection*{Proof of Proposition \ref{prop4}}
We want to check whether equality between asymptotic risk and the expected PPV holds. Using Proposition \ref{prop3}: In the highly overparameterized regime, we demand equality
\begin{align*}
    \mathcal{R}_{wide} = \frac{F_1^2\psi_2 + (F_*^2+\tau^2)\omega_2^2}{\psi_2 - 2\omega_2\psi_2 + \omega_2^2\psi_2 - \omega_2^2} &\stackrel{*}{=} \frac{F_1^2}{1-\omega_2}.\\
    \iff \omega_2^3 + (\rho\psi_2-\rho-1)\omega_2^2 - \psi_2\rho\omega_2 &\stackrel{*}{=} 0.
\end{align*}
By Proposition 13.1, \cite{mei2022generalization}, this equation is met precisely by choosing $\bar{\lambda}\equiv\lambda^{opt}$, only possible if $\rho < \rho^*$ per the definition of Proposition \ref{prop4}. Otherwise (i.e., if $\rho > \rho^*$), the equality is unsatisfiable, and furthermore, one can use the property of the map $\omega\equiv\omega(\bar{\lambda},\psi,\zeta)$ in Lemma 13.1, \cite{mei2022generalization}, to show that $\omega_2^3 + (\rho\psi_2-\rho_1)\omega_2^2 - \psi_2\rho\omega_2 > 0$.

In the large sample regime, we similarly want to check the equality
\begin{align*}
    \frac{\omega_1^2/\zeta^2+\psi_1}{\psi_1-2\psi_2\omega_1 + \psi_1\omega_1^2-\omega_1^2} &\stackrel{*}{=} \frac{1}{1-\omega_1^2}.\\
    \iff \omega_1^3 + (\zeta^2\psi_1-\zeta^2-1)\omega_1^2 - \psi_1\zeta^2\omega_1 &\stackrel{*}{=} 0.
\end{align*}
In this case, the demanded equality does not have $\rho$ in it. Utilizing symmetry and the same argument as in Proposition 13.1, \cite{mei2022generalization}, one can show that the absence of $\rho$ implies the equality is always satisfiable by choosing $\bar{\lambda} = 0$, which also corresponds to the minimizer of the asymptotic risk: ${\arg\min}_{\bar{\lambda}\geq 0}\mathcal{R}_{lsamp} = 0$.

\end{document}